\documentclass[a4paper,fleqn]{cas-dc}

\usepackage[authoryear,longnamesfirst]{natbib}
\usepackage{xurl}
\begin{document}
\let\WriteBookmarks\relax
\def\floatpagepagefraction{1}
\def\textpagefraction{.001}
\ExplSyntaxOn
\cs_gset:Npn \__first_footerline: { }
\cs_gset:Npn \__cas_foot:
  {
    \parbox[t]{\textwidth}
      {
        \rule{\textwidth}{.2pt}\\
        \hfill Page~\thepage{}~of~\lastpage
      }
  }
\cs_gset:Npn \__first_foot:
  {
    \parbox[t]{\textwidth}
      {
        \rule{\textwidth}{.2pt}\\
        \hfill Page~\thepage{}~of~\lastpage
      }
  }
\ExplSyntaxOff

\shorttitle{Community-aware open-set plankton recognition}
\shortauthors{Chen et al.}

\title[mode=title]{Community-aware evaluation and threshold calibration for open-set plankton image recognition}

\author[1]{Xi Chen}
\fnmark[1]

\affiliation[1]{
  organization={School of Computer Science and Cyber Engineering, Guangzhou University},
  city={Guangzhou},
  postcode={510006},
  country={China}
}

\author[2]{Eryuan Huang}
\fnmark[1]

\affiliation[2]{
  organization={School of Environment, South China Normal University},
  city={Guangzhou},
  postcode={510631},
  country={China}
}

\author[3]{Yingjun Xiao}

\affiliation[3]{
  organization={School of Artificial Intelligence, Guangzhou University},
  city={Guangzhou},
  postcode={510006},
  country={China}
}

\author[4]{Gang Fang}
\cormark[1]
\ead{gangf@gzhu.edu.cn}

\affiliation[4]{
  organization={Institute of Computing Science and Technology, Guangzhou University},
  city={Guangzhou},
  postcode={511370},
  country={China}
}

\fntext[1]{These authors contributed equally to this work.}
\cortext[1]{Corresponding author.}

\begin{abstract}
Automated plankton image recognition is increasingly used in aquatic ecosystem monitoring, but deployed classifiers inevitably encounter unseen taxa and non-target particles. Open-set recognition methods are usually evaluated with sample-level metrics such as AUROC, AUPR, and FPR@95\% unknown-recall operating points, whereas ecological monitoring depends on community-level estimates of taxon abundance and diversity. This study examines the mismatch between these objectives using controlled pseudo-communities and three datasets spanning marine zooplankton imaged by ZooScan, marine phytoplankton imaged by IFCB, and freshwater plankton imaged by an in-situ camera. We define Open-Set Community Distortion (OSCD), a Bray-Curtis-style error over known taxa plus an unknown bin, with directional components distinguishing known-taxon overestimation from underestimation. Closed-set classifiers achieved high known-class accuracy, but unknown samples were often absorbed with high confidence and in structured ways. Sample-level OOD metrics were not sufficient to select ecological operating points: for MSP, FPR@95\% unknown-recall thresholds produced large test-community OSCD on all three datasets mainly because true known taxa were over-rejected into the unknown bin. Community-aware threshold calibration reduced MSP OSCD relative to fixed 95\% known recall on SYKE-ZooScan 2024 and SYKE-IFCB 2022; on ZooLake the fixed-recall baseline was already close to the community-aware threshold, and the best community-level method was a prototype-distance variant rather than MSP. The benefit of community-aware calibration therefore depends on validation-community representativeness and the gap between fixed recall and the community optimum. These results show that open-set plankton recognition should be evaluated as an ecological measurement problem, not only as a sample-level detection task.
\end{abstract}

\begin{highlights}
\item Sample-level rejection thresholds can severely distort plankton community estimates across instruments and ecosystems.
\item Structured unknown absorption concentrates community errors in specific known taxa rather than distributing them uniformly.
\item OSCD directionality shows that aggressive rejection mainly underestimates known taxa; this pattern replicated on all three datasets.
\item Community-aware threshold calibration can improve OSCD when validation communities reflect deployment and the fixed-recall baseline leaves room for improvement.
\end{highlights}


\maketitle

\section{Introduction}
\label{sec:introduction}

Plankton communities are sensitive indicators of aquatic ecosystem state. Changes in their composition can reflect shifts in nutrient conditions, trophic interactions, harmful blooms, and broader environmental disturbance. Imaging systems such as ZooScan and flow-through imaging platforms have made it possible to collect plankton observations at a scale that cannot be handled by manual taxonomic annotation alone \citep{gorsky2010zooscan,sosik2007ifcb,irisson2022machine}. As a result, automated image classification has become an important component of plankton monitoring workflows.

The ecological use of these classifiers, however, is rarely limited to assigning a label to a single image. In monitoring practice, the quantities of interest are community-level summaries: relative abundances of taxa, diversity indices, evenness, richness, and the identity of dominant groups \citep{bray1957ordination,shannon1948mathematical,simpson1949measurement,pielou1966measurement}. A classifier that performs well on average at the image level may still produce biased ecological conclusions if its errors are structured. Systematic overestimation of one taxon can create a false apparent bloom, while systematic underestimation can make an abundant taxon appear depleted.

This distinction becomes especially important under open-set deployment, where incomplete knowledge of possible test classes is the central difficulty \citep{scheirer2013openset,geng2021opensetsurvey}. A classifier trained on a fixed set of known taxa will inevitably encounter unseen target organisms and non-target particles such as bubbles, fibres, skins, beads, or debris. Under a closed-set decision rule, every such image must be forced into one of the known categories. This forced assignment can be highly confident rather than uncertain. In our SYKE-ZooScan 2024 experiments, unknown samples reached a mean closed-set confidence of 0.71, and 28.0\% exceeded 0.90 confidence. More importantly, the absorption pattern was not uniform: specific unknown categories were consistently misclassified into specific known taxa, meaning that open-set errors translate into structured, taxon-specific biases in community abundance estimates rather than random noise.

Open-set recognition and out-of-distribution detection methods address this problem by assigning an unknown score to each sample \citep{hendrycks2017baseline,liang2018odin,liu2020energy,geng2021opensetsurvey}. They are usually evaluated with sample-level metrics such as AUROC, AUPR, FPR@95\% unknown-recall operating points, or detection F1. These metrics answer whether known and unknown images can be separated, but monitoring systems deploy a thresholded classifier whose outputs are then aggregated into communities. The unresolved question is therefore not only whether unknown images are detectable, but whether the chosen threshold preserves abundance estimates.

This paper studies thresholded open-set recognition as an abundance-estimation problem in plankton image recognition. We use a frozen DINOv2 image encoder with a linear known-class classifier, post-hoc OOD scores, and controlled pseudo-community experiments that isolate the effect of community structure. The framework is evaluated on three datasets: SYKE-ZooScan 2024 as the primary marine zooplankton benchmark, SYKE-IFCB 2022 as a cross-instrument marine phytoplankton validation, and ZooLake as a freshwater validation dataset with a long-tailed class distribution.

The contributions are organized around three parts of this evaluation problem:
\begin{itemize}
\item \textbf{Evaluation metric.} We define Open-Set Community Distortion (OSCD), a Bray-Curtis-style community error over known taxa plus an unknown bin, together with directional diagnostic components OSCD$^{+}$ and OSCD$^{-}$ that separate known-taxon overestimation from underestimation.
\item \textbf{Empirical diagnosis.} We show across three datasets that closed-set unknown absorption is structured by taxon, and that sample-level threshold criteria can distort abundance estimates when aggressive rejection removes true known taxa from the estimated community.
\item \textbf{Calibration protocol.} We evaluate threshold selection by validation-community OSCD and identify when this community-aware calibration improves test-community estimates, when a fixed-recall baseline is already close to optimal, and when validation communities do not transfer to deployment-like communities.
\end{itemize}

\section{Related work}
\label{sec:related-work}

Automated plankton imaging has developed from specialized sampling instruments into an important component of high-frequency ecological monitoring. Systems such as ZooScan, optical plankton samplers, FlowCAM, and Imaging FlowCytobot made it possible to collect and process plankton images at scales beyond manual microscopy \citep{davis2004realtime,gorsky2010zooscan,sosik2007ifcb,alvarez2014flowcam,campbell2013ifcb}. More recent work has coupled such instruments with automated image analysis and deep learning workflows for zooplankton and phytoplankton monitoring \citep{romagnan2016highfrequency,merz2021aquascope,kraft2022operational,irisson2022machine}. These studies establish the practical value of image-based plankton observation, but their evaluation is usually framed around classification accuracy, validation protocols, or operational throughput under a fixed label set \citep{gonzalez2017validation,ellen2024beyond}. They therefore do not fully address what happens when deployed classifiers encounter taxa or particles outside the training vocabulary.

Open-set recognition is directly concerned with this deployment condition \citep{scheirer2013openset,geng2021opensetsurvey}. In computer vision, classical and post-hoc approaches include OpenMax, maximum-softmax-probability scoring, ODIN, energy-based scoring, Mahalanobis feature distances, and related prototype or distance-based criteria \citep{bendale2016openmax,hendrycks2017baseline,liang2018odin,lee2018mahalanobis,liu2020energy}. These methods are typically compared using sample-level detection metrics such as AUROC, AUPR, FPR@95\% unknown-recall operating points, or detection F1. In plankton recognition, open-set evaluation has only recently been studied explicitly; for example, \citet{kareinen2024opensetplankton} evaluate unknown-class handling for plankton imagery. Such work is close in motivation to the present study, but it still treats open-set performance primarily as a sample-level recognition problem. It does not ask whether the selected rejection threshold preserves downstream ecological estimates of community composition.

The threshold-calibration problem studied here is also distinct from standard neural-network confidence calibration. Methods such as temperature scaling adjust predicted probabilities so that confidence better matches sample-level correctness likelihood \citep{guo2017calibration}. In contrast, community-aware calibration in this study leaves the classifier and OOD scores fixed, and selects an operating threshold to minimize validation-community abundance distortion. The calibration target is therefore an ecological measurement error, not probability calibration of individual predictions.

Ecological monitoring, by contrast, is built around aggregate community quantities. Relative abundance, Bray-Curtis dissimilarity, Shannon diversity, Simpson diversity, evenness, richness, and dominant-taxon structure are routinely used to describe ecological state and compare plankton communities \citep{bray1957ordination,shannon1948mathematical,simpson1949measurement,pielou1966measurement,pierella2022coupling}. In automated plankton workflows, this means that classifier outputs are not only image labels; they are inputs to ecological measurement. Prior validation studies have recognized that automated classification errors can affect abundance estimates and community summaries \citep{alvarez2014flowcam,gonzalez2017validation,romagnan2016highfrequency}. However, this literature generally assumes a closed set of target categories or evaluates aggregate agreement after classification. It does not systematically analyze how open-set rejection errors propagate into known-taxon overestimation, known-taxon underestimation, or threshold-dependent community distortion.

What remains missing is an evaluation that treats open-set plankton recognition as an ecological measurement problem: the operating threshold should be judged not only by whether it separates known and unknown images, but by how it changes the inferred community.

\section{Methods}
\label{sec:methods}

\begin{figure*}[t]
\centering
\includegraphics[width=\textwidth]{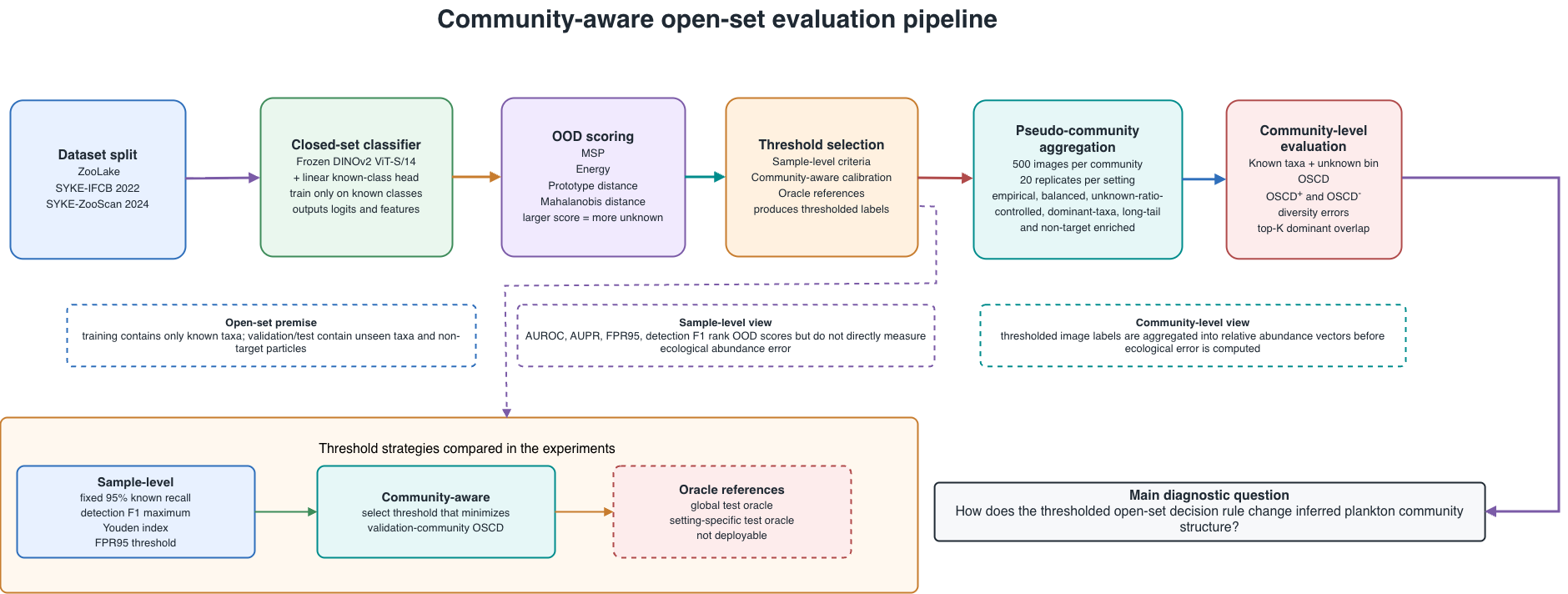}
\caption{Overview of the community-aware open-set evaluation pipeline. Each dataset is split into known classes and held-out unknown categories, a closed-set classifier is trained only on known classes, and post-hoc OOD scores are converted into thresholded labels using sample-level criteria, community-aware calibration, or oracle references. Thresholded labels are aggregated into pseudo-communities and evaluated as relative abundance vectors with OSCD, OSCD$^{+}$, OSCD$^{-}$, diversity errors, and dominant-taxon overlap.}
\label{fig:workflow}
\end{figure*}

The overall experimental workflow is summarized in Fig.~\ref{fig:workflow}. The experiments were designed to answer three questions: (RQ1) do closed-set models absorb unknown plankton categories in a structured way; (RQ2) do sample-level OOD scores and thresholds preserve community composition after aggregation; and (RQ3) when does selecting the threshold by validation-community OSCD improve test-community estimates? The following subsections describe the data split, pseudo-community construction, classifier and OOD scores, community-level metrics, and threshold-selection strategies used to answer these questions.

\subsection{Datasets and known/unknown split}
\label{subsec:dataset-split}

All experiments follow a common split design intended to emulate open-set deployment: a classifier is trained on a restricted set of known classes, then evaluated on images that also contain unseen target categories and non-target particles. Known classes are selected to represent a realistic monitoring vocabulary for the corresponding ecosystem. Unknown categories are defined as all remaining annotated categories, grouped into target unknowns and non-target unknowns based on whether they represent organisms of ecological interest or imaging artefacts and other non-target material. This two-way grouping is operationally defined and applies uniformly across datasets; it does not rely on assumptions about morphological or taxonomic distance between known and unknown classes.

SYKE-ZooScan 2024 serves as the primary benchmark. It contains 22,753 ZooScan images from 20 annotated categories collected from the Baltic Sea as part of the Open-Set Plankton Recognition dataset \citep{kareinen2024opensetplankton}. Ten categories were used as known classes for closed-set training: \path{Ceriodaphnia_sp}, \path{Copepoda_calanoida}, \path{Copepoda_cyclopoida}, \path{Copepoda_nauplius}, \path{Daphnia_sp}, \path{Evadne_sp}, \path{Gastropoda}, \path{Podon_sp}, \path{Polychaeta}, and \path{Synchaeta_sp}. For these categories, the official train, validation, and test image lists were preserved and mapped to the experimental splits. The remaining categories were assigned to validation unknown, test unknown, or non-target unknown roles. \path{Bubbles} and \path{Fibers_etc} constitute the non-target unknown group; all other held-out categories are target unknowns. This dataset provides the main experimental results, including the full pseudo-community analysis and applicability-boundary evaluation.

SYKE-IFCB 2022 provides cross-instrument and cross-taxon validation. It contains 63,074 IFCB images of phytoplankton from 50 categories collected from the Baltic Sea \citep{kraft2022operational}. Forty-one categories were used as known classes, eight categories were held out as target unknowns, and \path{Beads} was used as the non-target unknown category. The split was constructed manually to keep validation-unknown and test-unknown categories disjoint while using the same target/non-target logic as the primary benchmark. This dataset tests whether threshold-objective mismatch and community-aware calibration findings replicate on a different imaging platform and a different plankton group within the same broad geographic region.

ZooLake provides cross-ecosystem validation in a freshwater setting. It contains 17,940 usable images from 35 categories captured by an in-situ camera deployed at Lake Greifensee, Switzerland \citep{merz2021aquascope}. Nineteen categories were used as known classes, ten categories were held out as target unknowns, and six categories were treated as non-target unknowns. Because ZooLake categories are not strict taxonomic ranks, the target/non-target grouping is applied based on whether a category represents a planktonic organism or a non-living particle, skin, fish image, or ambiguous non-target label. Its freshwater community structure, different instrument type, and long-tailed class distribution test the generality and boundaries of the framework beyond the marine benchmarks.

\begin{table*}[t]
\caption{Dataset split overview. Target unknowns are held-out organism categories; non-target unknowns are imaging artefacts, particles, skins, beads, fish images, or ambiguous non-target labels depending on the dataset.}
\label{tab:dataset-overview}
\begin{tabular*}{\textwidth}{@{\extracolsep{\fill}}lrrrrrrl@{}}
\toprule
Dataset & Images & Classes & Known cls. & Target unk. cls. & Non-target cls. & Known imgs. & Role in study \\
\midrule
SYKE-ZooScan 2024 & 22,753 & 20 & 10 & 8 & 2 & 15,091 & Primary benchmark \\
SYKE-IFCB 2022 & 63,074 & 50 & 41 & 8 & 1 & 49,246 & Cross-instrument validation \\
ZooLake & 17,940 & 35 & 19 & 10 & 6 & 15,275 & Freshwater validation \\
\bottomrule
\end{tabular*}
\end{table*}

\subsection{Pseudo-community construction}
\label{subsec:pseudo-community-construction}

The community-level experiments use pseudo-communities generated by resampling individual images from the validation and test pools. This design deliberately keeps the image source, classifier, and labeling scheme fixed within each dataset while varying only the community composition. As a result, differences in community-level error can be attributed to controlled changes in abundance structure rather than to changes in model architecture or thresholding procedure.

Each pseudo-community contains 500 images sampled with replacement. For each eligible community setting, 20 independent communities were generated with random seed 42 for the main figures and detailed threshold analyses. Controlled unknown ratios were 0, 0.1, 0.2, and 0.4. Empirical communities were generated separately from the validation and test pools by sampling according to the observed split-specific image frequencies, and therefore used the observed unknown ratio of each split rather than a fixed ratio.

To check that the main community-level conclusions were not artefacts of a single pseudo-community draw, we repeated pseudo-community generation and threshold evaluation with five seeds (42--46), keeping the trained classifier, sample-level predictions, and OOD scores fixed. Cross-dataset OSCD summaries and applicability-boundary claims are reported as mean $\pm$ standard deviation across these five seeds where indicated. For applicability-boundary comparisons, paired $t$ tests compare the best sample-level strategy with the best community-aware strategy across matched seeds for each community setting; these tests are used as robustness diagnostics for resampling variation, not as evidence from independent datasets.

The shared community structures were empirical, balanced, unknown-ratio-controlled, dominant-taxa, long-tail, and non-target-enriched. Empirical communities preserve the observed image frequency of the corresponding split. Balanced communities allocate images approximately uniformly across known and unknown categories after setting the requested unknown ratio. Unknown-ratio-controlled communities fix the total unknown proportion but sample within known and unknown pools according to empirical frequencies. Dominant-taxa communities allocate 75\% of the known portion to the most frequent known class in the corresponding split. Long-tail communities use inverse class-frequency weights within the known and unknown portions, increasing the chance that rare categories appear. Non-target-enriched communities draw the unknown portion from non-target particles when such samples are available in the split.

The primary SYKE-ZooScan benchmark is used for detailed applicability-boundary analysis. The SYKE-IFCB and ZooLake datasets are used to test whether the main findings replicate under different instruments, plankton groups, and ecosystem structures. This division keeps the main paper focused while still using the external datasets to challenge the generality of the conclusions.

\subsection{Closed-set classifier and OOD scores}
\label{subsec:model-ood-scores}

For each dataset, the baseline classifier consists of a frozen DINOv2 ViT-S/14 image encoder \citep{oquab2023dinov2} followed by a trainable linear classification head over that dataset's known classes. Images were resized to $224 \times 224$ pixels and normalized with ImageNet channel statistics. During training, random horizontal and vertical flips were applied. The encoder weights were not updated; only the linear head was optimized. This choice reduces the number of experimental variables and focuses the study on how post-hoc open-set scores and threshold choices affect ecological community estimates.

The linear head was trained for 20 epochs using AdamW \citep{loshchilov2019adamw}, learning rate $10^{-3}$, weight decay $10^{-4}$, batch size 64, class-weighted cross-entropy, and automatic mixed precision when CUDA was available. The checkpoint with the highest known-class validation accuracy was used for all subsequent prediction and OOD scoring.

For an image $x$, the classifier produces logits $z(x)$ over known classes and a feature vector $f(x)$ from the frozen encoder. All OOD scores were defined so that larger values indicate stronger evidence that the sample is unknown. The maximum softmax probability score \citep{hendrycks2017baseline} is
\begin{equation}
s_{\mathrm{MSP}}(x) = 1 - \max_k \mathrm{softmax}_k(z(x)).
\end{equation}
The energy score \citep{liu2020energy} is
\begin{equation}
s_{\mathrm{Energy}}(x) = -T \log \sum_k \exp(z_k(x)/T),
\end{equation}
with temperature $T=1$.

For prototype-based scores, class prototypes were computed as the mean frozen-encoder feature of each known class in the training set. The raw prototype distance is the minimum Euclidean distance from $f(x)$ to any known class prototype. We additionally evaluated an L2-normalized Euclidean prototype distance and a cosine prototype distance. Mahalanobis scores \citep{lee2018mahalanobis} were computed as the minimum squared Mahalanobis distance to known class prototypes using a pooled covariance matrix estimated from centered known-class training features, with diagonal shrinkage applied before computing the pseudo-inverse.

\subsection{Community-level metrics}
\label{subsec:community-metrics}

For each community, individual classifier predictions were converted into a predicted abundance vector over $K$ known taxa plus one unknown bin. Let $p \in \mathbb{R}^{K+1}$ denote the true relative abundance vector and $\hat{p} \in \mathbb{R}^{K+1}$ denote the predicted relative abundance vector, both normalized to sum to one. The primary community-level error metric is Open-Set Community Distortion (OSCD), defined as Bray-Curtis distance \citep{bray1957ordination} on the known-plus-unknown abundance vectors:
\begin{equation}
\mathrm{OSCD}(p,\hat{p}) =
\frac{\sum_{i=1}^{K+1} |\hat{p}_i - p_i|}
{\sum_{i=1}^{K+1}(\hat{p}_i+p_i)}.
\end{equation}
Since both vectors are relative abundances, the denominator equals 2.

To diagnose the direction of known-taxon abundance errors, we computed two directional diagnostic components over known taxa:
\begin{equation}
\mathrm{OSCD}^{+} =
\sum_{i=1}^{K}\max(\hat{p}_i-p_i,0),
\end{equation}
\begin{equation}
\mathrm{OSCD}^{-} =
\sum_{i=1}^{K}\max(p_i-\hat{p}_i,0).
\end{equation}
Here $\mathrm{OSCD}^{+}$ measures overestimation of known taxa, which can occur when unknown samples are absorbed into known classes. In contrast, $\mathrm{OSCD}^{-}$ measures underestimation of known taxa, which can occur when true known samples are rejected into the unknown bin. These quantities are directional diagnostics rather than additive parts of OSCD. If $A=\mathrm{OSCD}^{+}$ and $B=\mathrm{OSCD}^{-}$, the unknown-bin difference is $B-A$ because both abundance vectors sum to one. Thus, for normalized abundance vectors, $\mathrm{OSCD}=\frac{1}{2}(A+B+|B-A|)=\max(A,B)$. This identity holds at the level of a single community. In practice, all reported OSCD, OSCD$^{+}$, and OSCD$^{-}$ values are means over pseudo-communities, so this single-community identity does not hold for the tabulated values; the directional components should be read as diagnostics rather than as an additive decomposition of mean OSCD. The two error directions have different ecological interpretations: the former can create false apparent blooms of known taxa, whereas the latter can make real taxa appear depleted or absent.

OSCD was used as the primary composition-level endpoint. Additional community metrics were computed as secondary ecological diagnostics: mean absolute relative abundance error over known taxa, Shannon error \citep{shannon1948mathematical}, Simpson error \citep{simpson1949measurement}, Pielou evenness error \citep{pielou1966measurement}, richness error, and top-$K$ dominant taxa overlap with $K=3$.

\subsection{Threshold selection strategies}
\label{subsec:threshold-strategies}

For each OOD method, candidate thresholds were generated from score quantiles and evaluated by assigning samples with scores above the threshold to the unknown bin. The scan used a grid of 401 score quantiles plus extrema. At each threshold, sample-level OOD metrics and community-level metrics were computed on the corresponding validation and test subsets.

Threshold selection strategies were grouped into sample-level, community-aware, and oracle strategies. Sample-level strategies selected thresholds using only validation samples: fixed known recall 95\%, detection F1 maximum, Youden index maximum \citep{youden1950index}, and an FPR@95\% unknown-recall threshold when possible. For this threshold, the positive class is the unknown class, so the operating point is defined by high unknown recall rather than by known-class true-positive rate. The community-aware strategy selected the threshold that minimized mean validation-community OSCD. Oracle strategies were used only as theoretical upper bounds: the global oracle minimized mean OSCD over all test communities, whereas the setting oracle minimized mean OSCD within each test community type and unknown-ratio setting.

For the ecological-objective analysis, additional validation-community thresholds were selected with the same threshold grid but different secondary objectives: minimizing mean relative abundance error, Shannon error, Simpson error, Pielou error, or richness error, and maximizing mean top-3 dominant-taxon overlap. These objective-aware thresholds were used to diagnose trade-offs among ecological endpoints rather than to replace OSCD as the primary composition metric.

\begin{table*}[t]
\caption{Threshold selection strategies used in the experiments. Oracle strategies are not deployable and are included only as theoretical reference points.}
\label{tab:threshold-strategies}
\begin{tabular*}{\textwidth}{@{\extracolsep{\fill}}lll@{}}
\toprule
Strategy & Selection criterion & Data used for selection \\
\midrule
Fixed known recall 95\% & Known recall closest to 0.95 & Validation samples \\
Detection F1 max & Maximum unknown-detection F1 & Validation samples \\
Youden index max & Maximum known recall + unknown rejection - 1 & Validation samples \\
FPR@95\% unknown recall & Threshold targeting 95\% unknown rejection & Validation samples \\
Community-aware & Minimum mean OSCD & Validation communities \\
Global oracle & Minimum mean OSCD over all test communities & Test communities \\
Setting oracle & Minimum mean OSCD within each test setting & Test communities by setting \\
\bottomrule
\end{tabular*}
\end{table*}

\section{Results}
\label{sec:results}

\subsection{Closed-set models absorb unknown samples with high confidence}
\label{subsec:closed-set-absorption}

The frozen DINOv2 encoder with a linear classification head provided strong closed-set baselines on all datasets. Known-class validation/test accuracy was 0.9655 on SYKE-ZooScan 2024, 0.9643 on SYKE-IFCB 2022, and 0.9760 on ZooLake. The subsequent open-set errors therefore do not reflect failed closed-set training.

However, unknown samples were not reliably assigned low confidence by the closed-set classifier (Table~\ref{tab:cross-confidence}). Across all unknown samples, mean maximum softmax confidence was 0.7119 on SYKE-ZooScan 2024, 0.8234 on SYKE-IFCB 2022, and 0.7717 on ZooLake. The fraction of unknown images exceeding 0.90 confidence was 28.0\%, 50.4\%, and 38.5\%, respectively. Non-target unknowns were also frequently absorbed with high confidence, especially on SYKE-ZooScan 2024 and SYKE-IFCB 2022.

\begin{table*}[t]
\caption{Closed-set confidence statistics across datasets. Accuracy is reported for known samples; $c$ denotes maximum softmax confidence. Target unknowns are held-out organism categories, while non-target unknowns are artefacts, particles, skins, beads, fish images, or ambiguous non-target labels depending on the dataset.}
\label{tab:cross-confidence}
\begin{tabular*}{\textwidth}{@{\extracolsep{\fill}}llrrrr@{}}
\toprule
Dataset & Group & $n$ & Acc. & Mean $c$ & $c \geq 0.90$ \\
\midrule
SYKE-ZooScan 2024 & Known & 6030 & 0.9655 & 0.9620 & 0.8922 \\
SYKE-ZooScan 2024 & Target unknown & 6826 & -- & 0.6934 & 0.2439 \\
SYKE-ZooScan 2024 & Non-target unknown & 836 & -- & 0.8624 & 0.5706 \\
SYKE-IFCB 2022 & Known & 19682 & 0.9643 & 0.9697 & 0.9139 \\
SYKE-IFCB 2022 & Target unknown & 13703 & -- & 0.8228 & 0.5032 \\
SYKE-IFCB 2022 & Non-target unknown & 125 & -- & 0.8871 & 0.6240 \\
ZooLake & Known & 4575 & 0.9760 & 0.9704 & 0.9163 \\
ZooLake & Target unknown & 1917 & -- & 0.7790 & 0.3881 \\
ZooLake & Non-target unknown & 748 & -- & 0.7531 & 0.3770 \\
\bottomrule
\end{tabular*}
\end{table*}

\begin{figure*}[t]
\centering
\includegraphics[width=\textwidth]{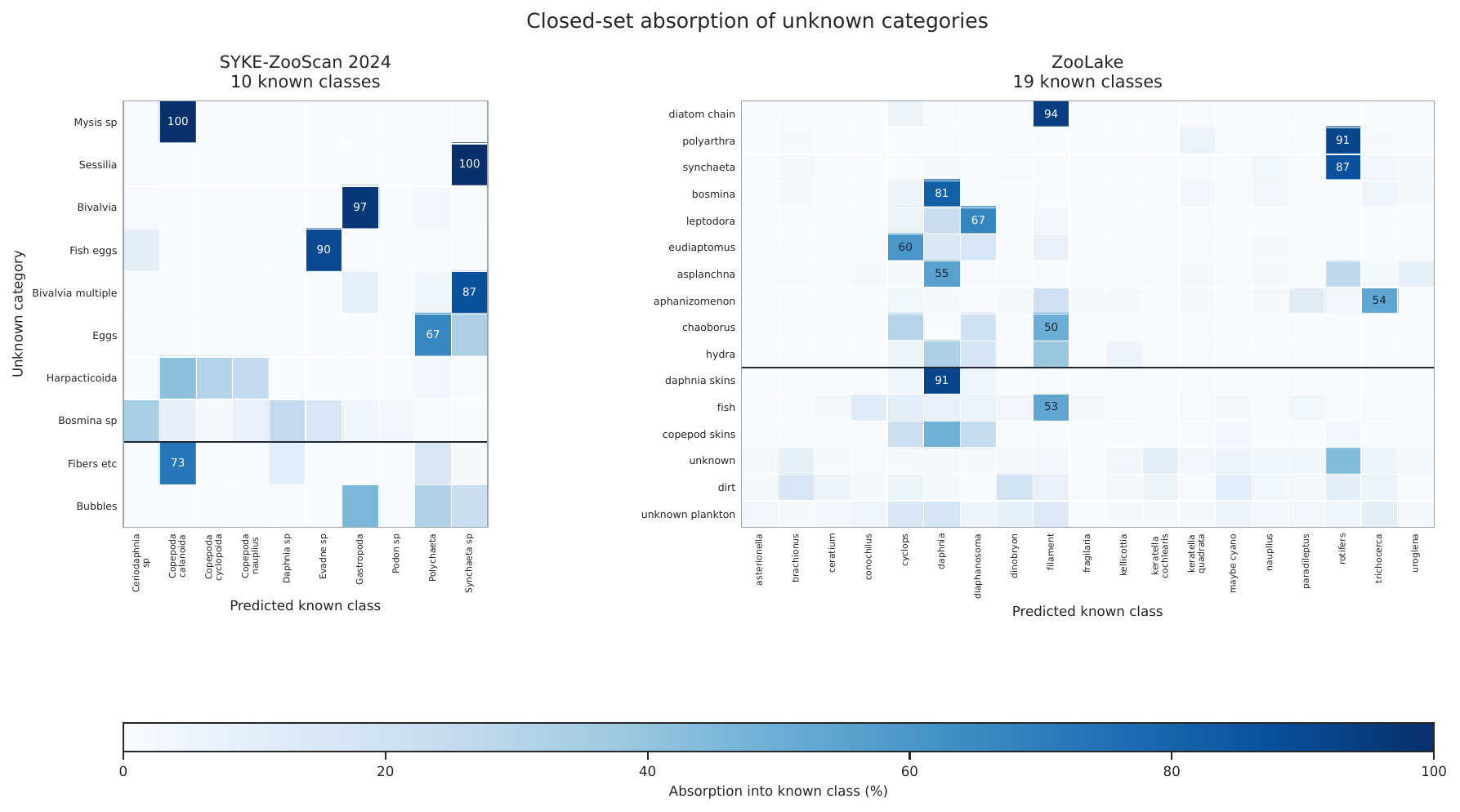}
\caption{Closed-set absorption of unknown categories into known classes on SYKE-ZooScan 2024 and ZooLake. Each panel corresponds to one dataset, each row is an unknown category, and each column is a known training category. Cell color shows the percentage of samples in that unknown category predicted as the corresponding known class by the closed-set classifier; annotated cells mark high-concentration absorption. Rows are ordered with target unknowns first and non-target unknowns below the horizontal divider. The concentration of mass in specific columns shows that unknown samples are absorbed in structured, category-dependent ways rather than being uniformly distributed across known taxa.}
\label{fig:unknown-absorption-main}
\end{figure*}

\begin{figure*}[t]
\centering
\includegraphics[width=\textwidth]{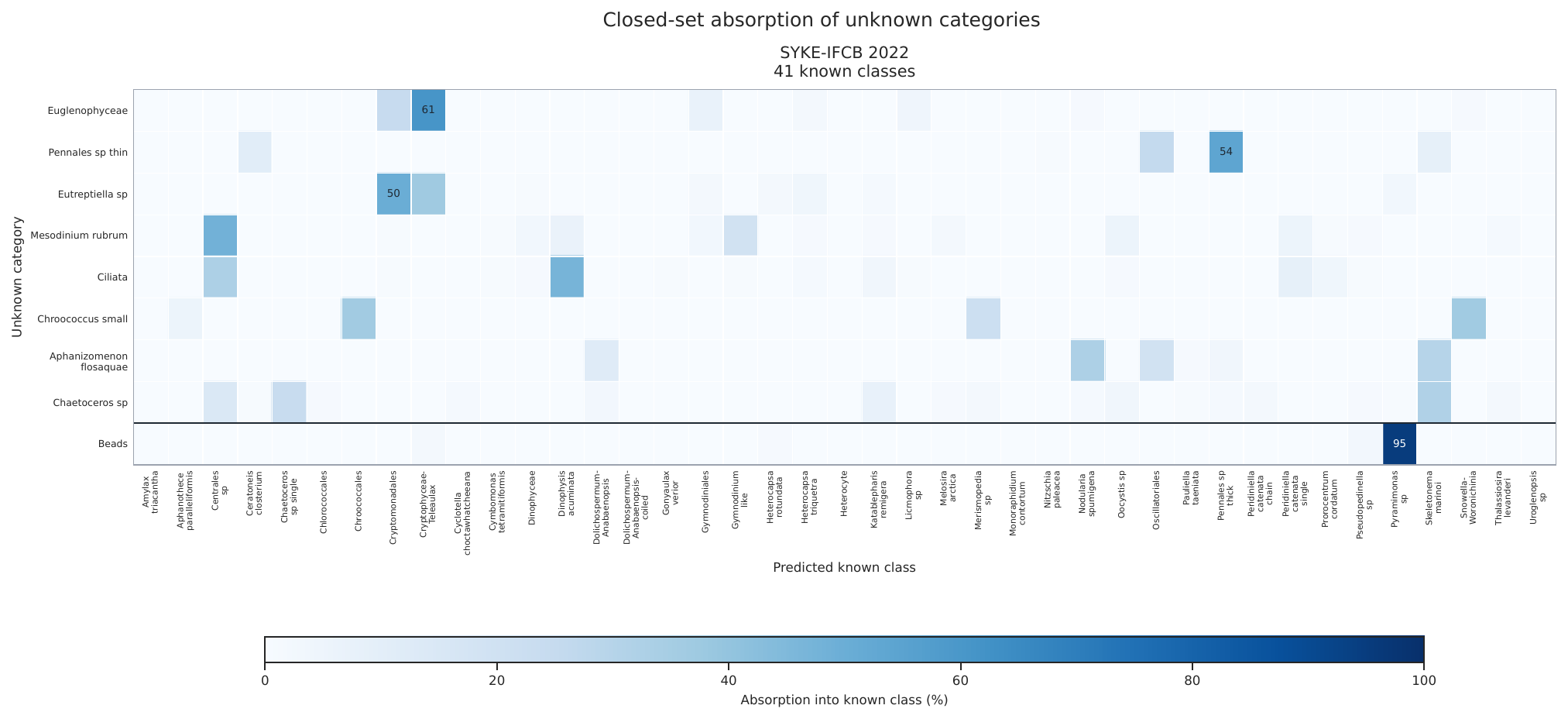}
\caption{Closed-set absorption of unknown categories into known classes on SYKE-IFCB 2022. The IFCB dataset has 41 known classes, so it is shown separately to keep labels legible while preserving the same color scale and annotation style as Fig.~\ref{fig:unknown-absorption-main}. Rows are ordered with target unknowns first and non-target unknowns below the horizontal divider. The same row-wise concentration pattern appears despite the larger known-class vocabulary and different imaging instrument.}
\label{fig:unknown-absorption-ifcb}
\end{figure*}

More ecologically consequential than the average confidence level is the structure of the absorption pattern. The absorption matrices in Figs.~\ref{fig:unknown-absorption-main} and \ref{fig:unknown-absorption-ifcb} show that unknown categories were not misclassified uniformly across known taxa. Instead, each unknown category concentrated its errors in one or a few specific known classes. On SYKE-ZooScan 2024, for example, \path{Bivalvia} was predominantly absorbed into \path{Gastropoda}, while \path{Fibers_etc} was often absorbed into \path{Copepoda_calanoida}. Similar row-wise concentration appears on SYKE-IFCB 2022 and ZooLake, although the specific receiving taxa differ by dataset. This structured absorption means that open-set errors do not average out at the community level. A sample dominated by \path{Bivalvia} will systematically overestimate \path{Gastropoda} abundance, not produce a uniform elevation across all known taxa. This mechanism motivates community-level evaluation: the ecological damage of open-set errors depends not only on how many unknown samples are absorbed, but also on which known taxa receive that absorbed abundance.

\subsection{Sample-level OOD scores provide an incomplete ranking}
\label{subsec:sample-ood-results}

We next evaluated whether post-hoc OOD scores could separate known from unknown samples before considering community-level consequences. MSP was the strongest test-split AUROC method on all three datasets, reaching 0.9164 on SYKE-ZooScan 2024, 0.8828 on SYKE-IFCB 2022, and 0.8932 on ZooLake (Table~\ref{tab:cross-ood}). This confirms that a simple confidence-based score remains a strong sample-level baseline across instruments and ecosystems.

\begin{table*}[t]
\caption{Cross-dataset sample-level OOD and community-ranking summary. Test AUROC reports the best sample-level method on each dataset. FPR@95\% unknown recall is the false-positive rate at the operating point targeting 95\% unknown rejection. Rank consistency compares the test AUROC ranking with the test-community OSCD ranking obtained after validation-community threshold selection; larger Spearman values indicate closer agreement.}
\label{tab:cross-ood}
\begin{tabular*}{\textwidth}{@{\extracolsep{\fill}}llrrrll@{}}
\toprule
Dataset & Best AUROC method & Test AUROC & FPR@95\% unk. recall & Spearman & Sample top & Community top \\
\midrule
SYKE-ZooScan 2024 & MSP & 0.9164 & 0.3067 & 0.9048 & MSP & MSP \\
SYKE-IFCB 2022 & MSP & 0.8828 & 0.3619 & 0.2857 & MSP & MSP \\
ZooLake & MSP & 0.8932 & 0.3785 & 0.3810 & MSP & Prototype distance \\
\bottomrule
\end{tabular*}
\end{table*}

The relationship between sample-level and community-level rankings was not stable across datasets. SYKE-ZooScan 2024 showed relatively high rank agreement between AUROC and OSCD. In SYKE-IFCB 2022, MSP remained the top method under both summaries, but the overall method-ranking agreement was weak (Spearman $r=0.2857$). ZooLake exposed a more direct method-level mismatch: MSP had the highest sample-level AUROC, but raw prototype distance produced the best community-level OSCD after validation-community threshold selection. The large cross-dataset variation in rank consistency indicates that method ordering depends on the interaction between score geometry, class structure, and community composition. Sample-level OOD quality is therefore useful for screening methods, but it is not a reliable substitute for evaluating the thresholded classifier as a community estimator. The more consequential mismatch is not only which score ranks highest, but which threshold is selected for deployment.

\subsection{Sample-level optimal thresholds do not align with community-level optima}
\label{subsec:threshold-mismatch-results}

This appears most clearly at the threshold-selection level. Closed-set or weakly rejected predictions expose one ecological error direction: unknown categories can be absorbed into known taxa, increasing OSCD$^{+}$. Overly aggressive rejection exposes the opposite direction: true known samples can be assigned to the unknown bin, increasing OSCD$^{-}$. Threshold selection determines which of these two abundance errors dominates.

\begin{figure*}[t]
\centering
\includegraphics[width=\textwidth]{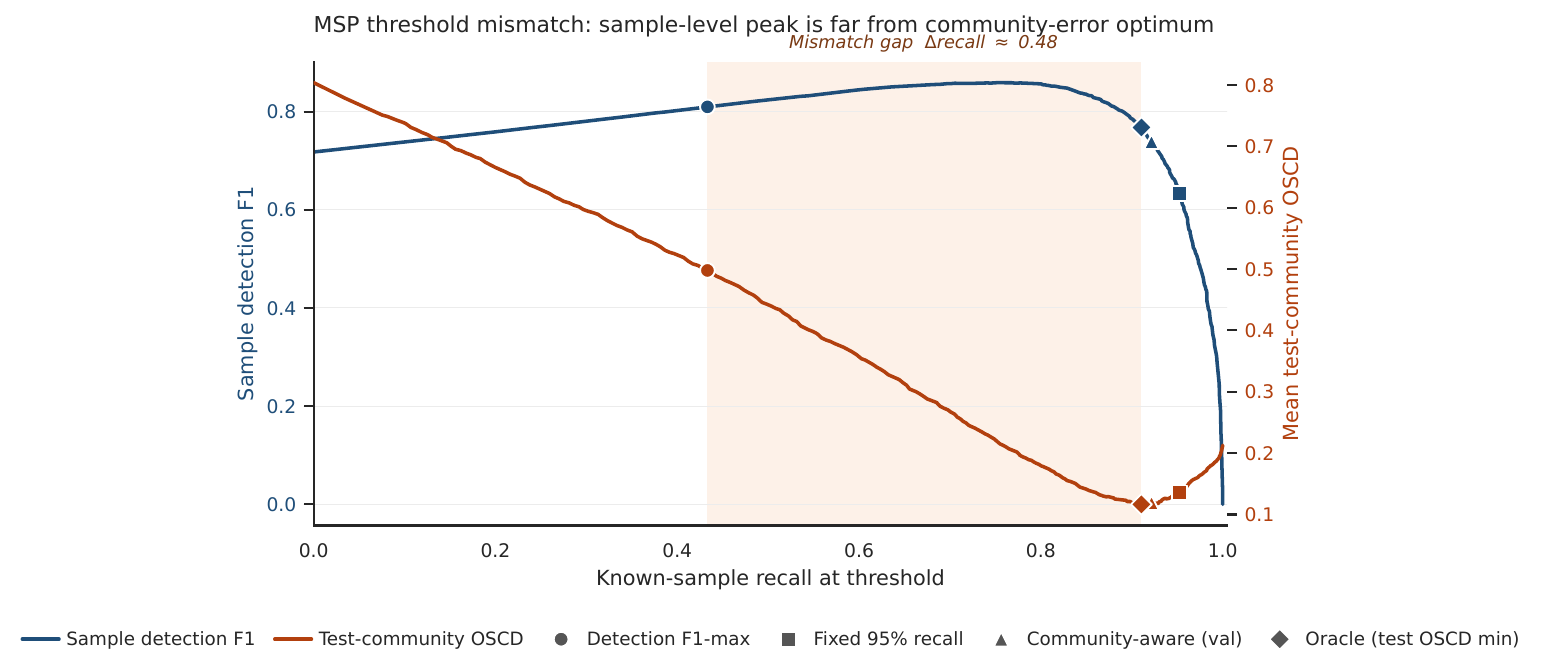}
\caption{MSP threshold mismatch between sample-level detection and community-level error on SYKE-ZooScan 2024. The blue curve (left axis) shows sample detection F1 and the orange curve (right axis) shows mean test-community OSCD, both plotted as a function of known-sample recall at threshold; lower OSCD indicates better community estimation, while higher F1 indicates better sample-level detection. Markers indicate four representative threshold strategies. The shaded region highlights the recall gap between the sample-level F1 maximum and the community-level OSCD minimum ($\Delta$recall $\approx$ 0.48), illustrating why a threshold selected for image-level unknown detection can distort community composition.}
\label{fig:threshold-mismatch}
\end{figure*}

For MSP on the primary SYKE-ZooScan benchmark, the sample-level detection-F1 optimum occurred at a much lower known-sample recall than the community-aware and oracle thresholds, whereas community OSCD continued to improve under less aggressive rejection (Fig.~\ref{fig:threshold-mismatch}). This did not merely change sample-level precision and recall; it changed the estimated community composition by moving true known individuals into the unknown bin. The community-aware threshold did not reduce OSCD by sacrificing known-sample retention: compared with the detection-F1 threshold, it increased test known-sample recall from 0.4275 to 0.9217 while reducing mean test-community OSCD from 0.4977 to 0.1191.

The OSCD directionality made this failure mode explicit. On all three datasets, MSP thresholds targeting 95\% unknown rejection produced large community errors dominated by OSCD$^{-}$, meaning that true known taxa were undercounted after being rejected into the unknown bin (Table~\ref{tab:cross-direction}). Thus, a sample-level criterion designed to reject unknowns can conflict with abundance estimation when it erases real known abundance.

\begin{table*}[t]
\caption{MSP threshold-objective mismatch across datasets. Values are reported for seed 42 to show the detailed directional decomposition for representative threshold strategies; cross-seed variation for community-level OSCD is reported in Table~\ref{tab:cross-ca}. OSCD$^{-}$ dominates the FPR@95\% unknown-recall rows, showing that the main error is underestimation of known taxa due to over-rejection.}
\label{tab:cross-direction}
\begin{tabular*}{\textwidth}{@{\extracolsep{\fill}}llrrr@{}}
\toprule
Dataset & Strategy & OSCD & OSCD$^{+}$ & OSCD$^{-}$ \\
\midrule
SYKE-ZooScan 2024 & Fixed95 & 0.1353 & 0.1220 & 0.0417 \\
SYKE-ZooScan 2024 & Detection F1 & 0.4977 & 0.0036 & 0.4977 \\
SYKE-ZooScan 2024 & FPR@95\% unk. recall & 0.6953 & 0.0000 & 0.6953 \\
SYKE-IFCB 2022 & Fixed95 & 0.1541 & 0.1351 & 0.0617 \\
SYKE-IFCB 2022 & Detection F1 & 0.2711 & 0.0084 & 0.2711 \\
SYKE-IFCB 2022 & FPR@95\% unk. recall & 0.4237 & 0.0021 & 0.4237 \\
ZooLake & Fixed95 & 0.1213 & 0.0995 & 0.0528 \\
ZooLake & Detection F1 & 0.1532 & 0.0333 & 0.1523 \\
ZooLake & FPR@95\% unk. recall & 0.5220 & 0.0023 & 0.5220 \\
\bottomrule
\end{tabular*}
\end{table*}

The main ecological cost of aggressive sample-level rejection was not unknown contamination of known taxa. Instead, aggressive rejection removed real known taxa from the known community, causing systematic underestimation of known abundance. This effect is invisible if evaluation stops at AUROC or detection F1, but becomes clear once the same thresholds are evaluated as community estimators.

The directional components should be read as diagnostics rather than additive summaries of the mean OSCD. For example, on SYKE-ZooScan 2024 under fixed 95\% known recall, mean OSCD was 0.1353 whereas $\max(\mathrm{mean}\ \mathrm{OSCD}^{+},\mathrm{mean}\ \mathrm{OSCD}^{-})$ was 0.1220. Under detection-F1 and FPR@95\% unknown-recall thresholds, however, the same comparison was essentially equal because OSCD$^{-}$ dominated the error. This numerical gap shows that the mean directional components preserve information about error asymmetry across communities even when the single-community identity $\mathrm{OSCD}=\max(\mathrm{OSCD}^{+},\mathrm{OSCD}^{-})$ does not hold after averaging.

\begin{figure*}[t]
\centering
\includegraphics[width=\textwidth]{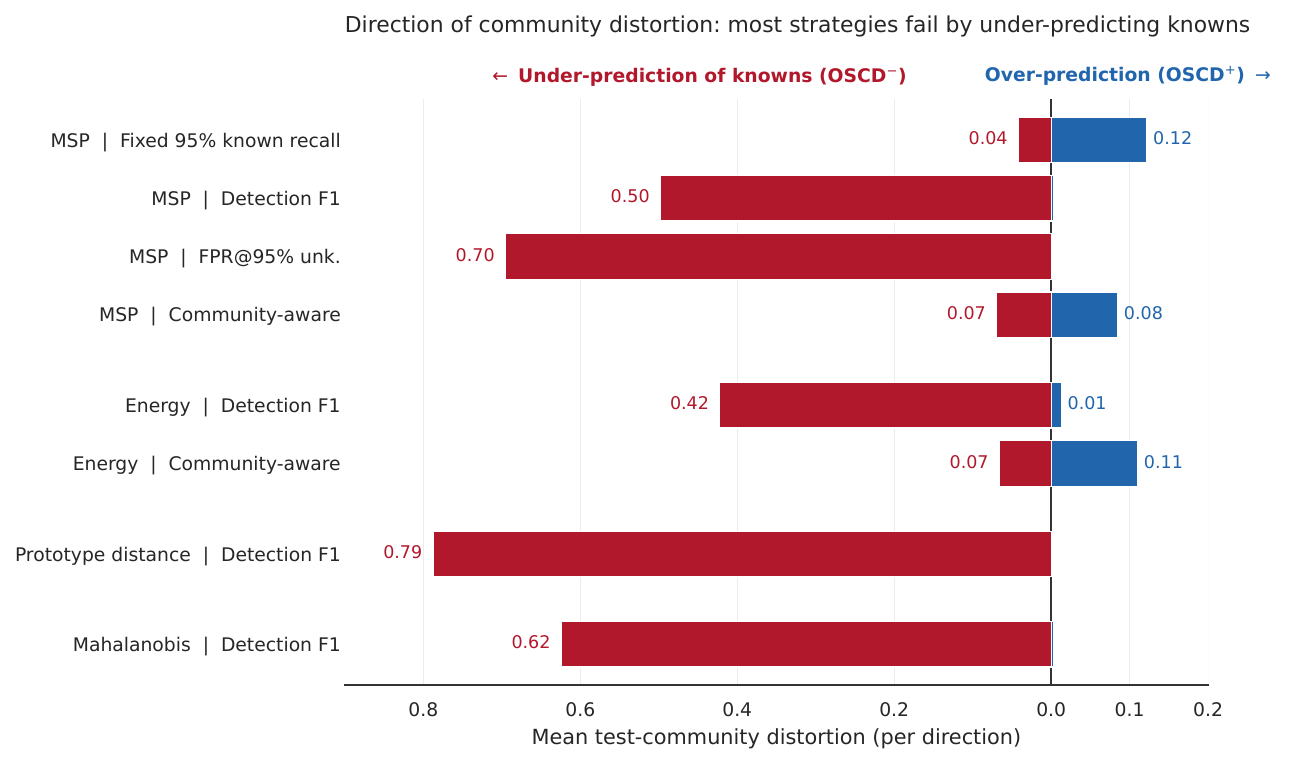}
\caption{Directional decomposition of community distortion under representative threshold strategies on SYKE-ZooScan 2024. Red bars extend leftward by mean OSCD$^{-}$, the under-prediction of known taxa, and blue bars extend rightward by mean OSCD$^{+}$, the over-prediction of known taxa. Aggressive sample-level thresholds such as detection-F1 and FPR@95\% unknown recall are dominated by OSCD$^{-}$, indicating that the primary ecological error is known-taxon underestimation caused by over-rejection. Table~\ref{tab:cross-direction} summarizes the same directional pattern across all three datasets.}
\label{fig:oscd-direction}
\end{figure*}

\subsection{Cross-dataset effect of community-aware calibration}
\label{subsec:community-aware-results}

Community-aware threshold calibration directly optimized the threshold for mean validation-community OSCD. Across five pseudo-community seeds, MSP fixed known recall 95\% achieved test OSCD $0.1353 \pm 0.0005$ on the primary SYKE-ZooScan benchmark, whereas community-aware calibration reduced OSCD to $0.1184 \pm 0.0007$, close to the global test oracle of $0.1162 \pm 0.0003$. The same pattern replicated on SYKE-IFCB 2022: MSP community-aware calibration reduced OSCD from $0.1540 \pm 0.0008$ under fixed 95\% known recall to $0.1416 \pm 0.0004$, with a global oracle of $0.1357 \pm 0.0007$ (Table~\ref{tab:cross-ca}).

\begin{table*}[t]
\caption{Cross-dataset community-level OSCD performance across five pseudo-community seeds, reported as mean $\pm$ standard deviation. Lower OSCD is better. The best non-oracle method may differ from MSP because method ranking can change when evaluated at the community level.}
\label{tab:cross-ca}
\footnotesize
\setlength{\tabcolsep}{3pt}
\begin{tabular*}{\textwidth}{@{\extracolsep{\fill}}llccccc@{}}
\toprule
Dataset & Best non-oracle method & Closed-set & Best OSCD & Oracle & MSP CA & MSP fixed95 \\
\midrule
SYKE-ZooScan 2024 & MSP & $0.2130 \pm 0.0003$ & $0.1184 \pm 0.0007$ & $0.1162 \pm 0.0003$ & $0.1184 \pm 0.0007$ & $0.1353 \pm 0.0005$ \\
SYKE-IFCB 2022 & MSP & $0.2154 \pm 0.0003$ & $0.1416 \pm 0.0004$ & $0.1357 \pm 0.0007$ & $0.1416 \pm 0.0004$ & $0.1540 \pm 0.0008$ \\
ZooLake & Prototype distance family & $0.1989 \pm 0.0002$ & $0.0907 \pm 0.0014$ & $0.0877 \pm 0.0007$ & $0.1201 \pm 0.0016$ & $0.1212 \pm 0.0006$ \\
\bottomrule
\end{tabular*}
\end{table*}

ZooLake defined the applicability boundary in two ways. For MSP, community-aware calibration and fixed 95\% known recall produced nearly identical OSCD ($0.1201 \pm 0.0016$ and $0.1212 \pm 0.0006$), indicating that fixed recall already sat close to the validation-community optimum. At the same time, the best community-level method was a prototype-distance variant rather than MSP, with best non-oracle OSCD $0.0907 \pm 0.0014$ close to the oracle value of $0.0877 \pm 0.0007$. Thus, threshold selection by validation-community OSCD improved MSP when the fixed-recall baseline left room for a better community-level operating point, while ZooLake changed the preferred OOD score more than the MSP threshold. Section~\ref{subsec:discussion-cross-dataset} returns to this distinction between score choice and threshold choice.

Because community-aware calibration uses OSCD as its selection objective, a natural follow-up question is whether OSCD itself behaves as a good proxy for the broader set of ecological summaries that monitoring programs report. The next section addresses this question.

\subsection{Ecological endpoint trade-offs beyond OSCD}
\label{subsec:ecological-summary-results}

The preceding results treat OSCD as the primary endpoint because it directly measures relative abundance distortion over known taxa plus an unknown bin. Ecological monitoring, however, also uses diversity, evenness, richness, and dominant-taxon summaries. We therefore examined whether OSCD behaves as a proxy for these secondary ecological diagnostics and whether thresholds calibrated for different ecological objectives select the same operating point. This analysis focuses on MSP on the primary SYKE-ZooScan benchmark, where the full ecological-objective scan was performed.

\begin{figure*}[t]
\centering
\includegraphics[width=\textwidth]{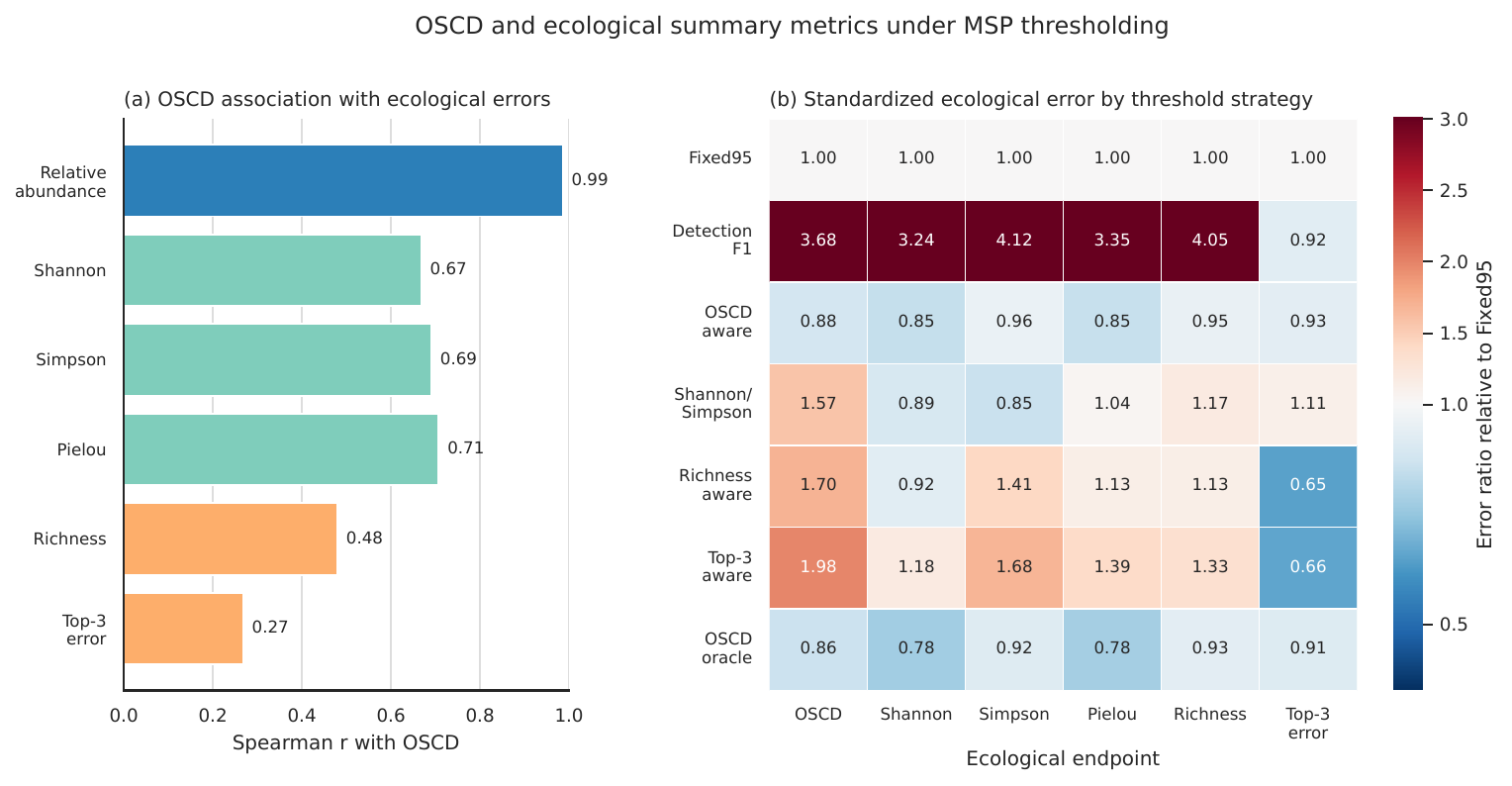}
\caption{Relationship between OSCD and secondary ecological diagnostics for MSP on SYKE-ZooScan 2024. (a) Spearman correlations between OSCD and ecological summary errors across test-community thresholds. Top-3 dominant-taxon overlap is converted to top-3 error as $1-\mathrm{overlap}$ so that larger values consistently indicate larger error. (b) Standardized ecological error of representative threshold strategies relative to fixed 95\% known recall. Values below 1 indicate improvement relative to fixed 95\% known recall, and values above 1 indicate degradation.}
\label{fig:ecological-tradeoffs}
\end{figure*}

Across MSP test-community thresholds, OSCD was almost identical to mean relative abundance error over known taxa, with Spearman $r=0.9856$ and Pearson $r=0.9929$ (Fig.~\ref{fig:ecological-tradeoffs}a). This confirms that OSCD is a strong abundance-composition endpoint. Its association with diversity-related summaries was weaker but still substantial: Spearman correlations were 0.6681 with Shannon error, 0.6910 with Simpson error, and 0.7072 with Pielou error. The relationship was weaker for richness error ($r=0.4799$) and weak for top-3 dominant-taxon error ($r=0.2681$). Thus, OSCD captures abundance distortion well, but it should not be interpreted as a universal surrogate for every ecological summary.

This distinction was also visible when comparing threshold strategies (Table~\ref{tab:ecological-objectives}). The validation detection-F1 threshold degraded several ecological summaries at once, producing OSCD 0.4977, Shannon error 0.6115, Simpson error 0.2386, Pielou error 0.2286, and richness error 2.1060. Compared with fixed 95\% known recall, OSCD-aware calibration reduced OSCD from 0.1353 to 0.1191. The same threshold also reduced Shannon error from 0.1887 to 0.1599, Simpson error from 0.0579 to 0.0554, Pielou error from 0.0682 to 0.0578, richness error from 0.5200 to 0.4940, and increased top-3 dominant-taxon overlap from 0.7193 to 0.7380.

\begin{table*}[t]
\caption{MSP threshold strategies evaluated with OSCD and secondary ecological diagnostics on SYKE-ZooScan 2024 test communities. Lower values are better except for top-3 overlap. Oracle rows are included only as theoretical references.}
\label{tab:ecological-objectives}
\footnotesize
\setlength{\tabcolsep}{3pt}
\begin{tabular*}{\textwidth}{@{\extracolsep{\fill}}lllrrrrrrr@{}}
\toprule
Strategy & Selected on & Target & Thr. & OSCD & Shannon & Simpson & Pielou & Richness & Top-3 \\
\midrule
Fixed95 & Sample val & Known recall & 276 & 0.1353 & 0.1887 & 0.0579 & 0.0682 & 0.5200 & 0.7193 \\
Detection-F1 & Sample val & Detection F1 & 72 & 0.4977 & 0.6115 & 0.2386 & 0.2286 & 2.1060 & 0.7407 \\
OSCD-aware & Val. comm. & OSCD & 239 & 0.1191 & 0.1599 & 0.0554 & 0.0578 & 0.4940 & 0.7380 \\
Shannon-/Simpson-aware & Val. comm. & Min Shannon or Simpson & 391 & 0.2124 & 0.1678 & 0.0495 & 0.0711 & 0.6100 & 0.6873 \\
Richness-aware & Val. comm. & Richness & 144 & 0.2294 & 0.1744 & 0.0817 & 0.0768 & 0.5860 & 0.8180 \\
Top-3-aware & Val. comm. & Top-3 overlap & 132 & 0.2672 & 0.2223 & 0.0974 & 0.0946 & 0.6900 & 0.8160 \\
Test OSCD oracle & Test comm. & OSCD & 227 & 0.1165 & 0.1465 & 0.0531 & 0.0530 & 0.4840 & 0.7440 \\
\bottomrule
\end{tabular*}
\end{table*}

Different ecological calibration targets selected different thresholds. Shannon-aware and Simpson-aware calibration selected a threshold with lower diversity error but higher OSCD. Richness-aware and top-3-overlap-aware calibration selected more aggressive thresholds and improved selected summaries while worsening abundance composition. These trade-offs show that there is not a single threshold that is optimal for all ecological endpoints. OSCD is a strong default when the monitoring target is relative abundance composition, but programs whose primary endpoints are diversity, evenness, or dominant-taxon structure should pair it with target-specific calibration.

\subsection{When does calibration transfer? Community-structure boundary analysis}
\label{subsec:boundary-results}

The applicability-boundary analysis was performed on the SYKE-ZooScan 2024 benchmark because its compact known-class vocabulary, explicit target/non-target unknown split, and primary-benchmark role made it best suited to controlled community-structure manipulation; cross-dataset replication of the main threshold-mismatch finding was already established in Sections~\ref{subsec:threshold-mismatch-results} and \ref{subsec:community-aware-results}. The analysis evaluated whether community-aware calibration is useful under different community structures and unknown ratios. The main boundary was an unknown-ratio transition around 0.2. In balanced, dominant-taxa, and unknown-ratio-controlled communities, community-aware calibration became useful once the unknown ratio reached 0.2, with the clearest and most stable gains at unknown ratio 0.4 (Table~\ref{tab:e5-boundaries}). For example, across five pseudo-community seeds, community-aware calibration reduced OSCD in balanced communities from $0.0990 \pm 0.0017$ to $0.0893 \pm 0.0013$ at unknown ratio 0.2, and from $0.1913 \pm 0.0020$ to $0.1741 \pm 0.0027$ at unknown ratio 0.4. In dominant-taxa communities, the corresponding improvements were from $0.0798 \pm 0.0016$ to $0.0648 \pm 0.0014$, and from $0.1512 \pm 0.0053$ to $0.1188 \pm 0.0039$.

\begin{table*}[t]
\caption{Selected applicability-boundary results on SYKE-ZooScan 2024 across five pseudo-community seeds, reported as mean $\pm$ standard deviation. Best sample OSCD is the best non-community-aware sample-level strategy for the setting. The paired $t$-test $p$ value compares best sample OSCD with community-aware OSCD across matched seeds and is reported as a robustness diagnostic, not as independent-dataset significance evidence. Lower OSCD is better.}
\label{tab:e5-boundaries}
\footnotesize
\setlength{\tabcolsep}{2.5pt}
\begin{tabular*}{\textwidth}{@{\extracolsep{\fill}}llccccl@{}}
\toprule
Community type & Unknown ratio & Best sample & CA & Setting oracle & $p$ & Recommendation \\
\midrule
Balanced & 0.2 & $0.0990 \pm 0.0017$ & $0.0893 \pm 0.0013$ & $0.0758 \pm 0.0011$ & $4.76{\times}10^{-6}$ & Boundary CA gain \\
Balanced & 0.4 & $0.1913 \pm 0.0020$ & $0.1741 \pm 0.0027$ & $0.0944 \pm 0.0027$ & $2.02{\times}10^{-5}$ & CA recommended \\
Dominant taxa & 0.2 & $0.0798 \pm 0.0016$ & $0.0648 \pm 0.0014$ & $0.0579 \pm 0.0009$ & $1.16{\times}10^{-6}$ & CA recommended \\
Dominant taxa & 0.4 & $0.1512 \pm 0.0053$ & $0.1188 \pm 0.0039$ & $0.0606 \pm 0.0026$ & $7.02{\times}10^{-5}$ & CA recommended \\
Non-target enriched & 0.1 & $0.0491 \pm 0.0015$ & $0.0252 \pm 0.0021$ & $0.0229 \pm 0.0010$ & $3.87{\times}10^{-5}$ & CA recommended \\
Non-target enriched & 0.2 & $0.0404 \pm 0.0020$ & $0.0515 \pm 0.0044$ & $0.0223 \pm 0.0016$ & $2.08{\times}10^{-3}$ & Sample preferred \\
Non-target enriched & 0.4 & $0.0316 \pm 0.0020$ & $0.0425 \pm 0.0025$ & $0.0206 \pm 0.0010$ & $9.90{\times}10^{-5}$ & Sample preferred \\
Long tail & 0.2 & $0.1574 \pm 0.0029$ & $0.1514 \pm 0.0021$ & $0.1444 \pm 0.0017$ & $1.77{\times}10^{-2}$ & Sample sufficient \\
Long tail & 0.4 & $0.2597 \pm 0.0058$ & $0.2564 \pm 0.0060$ & $0.1805 \pm 0.0035$ & $7.09{\times}10^{-5}$ & Sample sufficient \\
Unknown-ratio controlled & 0.2 & $0.0684 \pm 0.0020$ & $0.0552 \pm 0.0014$ & $0.0497 \pm 0.0018$ & $5.26{\times}10^{-5}$ & CA recommended \\
Unknown-ratio controlled & 0.4 & $0.1478 \pm 0.0033$ & $0.1104 \pm 0.0028$ & $0.0533 \pm 0.0026$ & $9.41{\times}10^{-6}$ & CA recommended \\
\bottomrule
\end{tabular*}
\end{table*}

Two boundary cases are particularly important. First, non-target-enriched communities did not follow a simple monotonic rule. Community-aware calibration helped at unknown ratio 0.1, but sample-level strategies were better at 0.2 and 0.4. This is consistent with the split design: non-target unknown categories were not represented in validation unknown communities with positive unknown ratios on the primary benchmark, so validation communities could not fully represent non-target-heavy deployment conditions.

Second, long-tail communities showed little improvement from OSCD-based community-aware calibration. At unknown ratio 0.4, the best sample-level strategy had OSCD $0.2597 \pm 0.0058$ and community-aware calibration had OSCD $0.2564 \pm 0.0060$, while the setting oracle was much lower at $0.1805 \pm 0.0035$. This suggests that long-tail settings may require threshold objectives that directly target diversity-sensitive metrics such as Shannon error, Pielou evenness, or richness error, rather than OSCD alone.

Across five pseudo-community seeds, the same trends were stable; the paired tests in Table~\ref{tab:e5-boundaries} are reported as robustness diagnostics rather than as independent-dataset significance evidence. The balanced community at unknown ratio 0.2 was the closest boundary: the mean gain was consistent ($0.0097 \pm 0.0006$, paired $p=4.76{\times}10^{-6}$), but the rule-based recommendation changed across seeds because the gain was near the recommendation threshold. The stronger CA-recommended settings were stable across all five seeds.

The empirical setting provided an additional check on validation representativeness. Empirical validation communities had mean unknown ratio 0.1004, whereas empirical test communities had mean unknown ratio 0.7061 and were dominated by a held-out organism category. In this case, community-aware thresholds selected from validation empirical communities did not transfer well to empirical test communities. This result clarifies the deployment assumption: validation communities must represent the expected deployment community structure for the calibrated threshold to be reliable.

\section{Discussion}
\label{sec:discussion}

\subsection{Cross-dataset generality and boundaries}
\label{subsec:discussion-cross-dataset}

Three points emerge consistently across the three datasets: closed-set absorption is structured and high-confidence; sample-level rejection thresholds primarily fail by underestimating known taxa; and the benefit of community-aware calibration is conditional on community structure and validation representativeness. The remainder of this section discusses each in turn and lays out the deployment assumptions and limitations that follow.

The three-dataset evaluation separates findings that were consistent from those that depended on dataset structure. Across marine zooplankton, marine phytoplankton, and freshwater plankton imagery, high closed-set accuracy did not prevent high-confidence unknown absorption, and sample-level threshold objectives could over-reject known organisms. The dataset-specific differences mattered for deployment. SYKE-IFCB 2022 had a higher unknown high-confidence rate than SYKE-ZooScan 2024, indicating that phytoplankton IFCB imagery can be at least as vulnerable to unknown absorption as the ZooScan benchmark.

ZooLake illustrates two distinct applicability dimensions of the framework. First, fixed 95\% known recall can already sit close to the community-aware MSP operating point, leaving little room for threshold-level calibration gain. Second, the best community-level OOD method was a prototype-distance variant rather than MSP, even though MSP had the highest sample-level AUROC. Together these show that community-level evaluation affects both where to threshold a score and which score to deploy, and that the size of the calibration gain depends on dataset structure.

\subsection{Why sample-level OOD metrics are insufficient for community estimation}
\label{subsec:discussion-sample-community}

AUROC, AUPR, and FPR@95\% unknown-recall operating points remain useful for screening broad method families, but they are less reliable as guides to the operating threshold. A monitoring system does not deploy an AUROC value; it deploys a thresholded decision rule. Sample-level detection metrics reward separating known and unknown images one by one, whereas community estimation depends on aggregate relative abundance. A threshold that improves unknown rejection can therefore damage the community estimate if it rejects many true known images into the unknown bin.

This distinction is especially important in plankton monitoring because many ecological decisions are based on abundance shifts rather than individual classifications. If a classifier rejects a large fraction of a dominant known taxon, the resulting community estimate may suggest a decline that did not occur. Conversely, if unknown particles are absorbed into a known class, the system may infer a bloom or compositional shift that is not present. Sample-level OOD metrics alone do not distinguish these two ecological failure modes.

\subsection{Ecological interpretation of OSCD directionality}
\label{subsec:discussion-oscd-direction}

The directional diagnostic components associated with OSCD make the threshold trade-off visible. OSCD$^{+}$ measures known-taxon overestimation. High OSCD$^{+}$ indicates that the predicted community contains too much abundance in one or more known taxa, which can happen when unknown taxa or non-target particles are absorbed into known classes. Such an error can create a false signal of taxon increase or bloom formation.

OSCD$^{-}$ measures known-taxon underestimation. High OSCD$^{-}$ indicates that real known taxa have been removed from the known portion of the predicted community, often because the OOD threshold is too aggressive. This error has a different ecological meaning: it can make common taxa appear rare, depleted, or absent. In the threshold experiments, the most damaging sample-level rejection strategies were dominated by OSCD$^{-}$ rather than OSCD$^{+}$. More aggressive unknown rejection does not simply protect the community from unknown contamination; it can erase known taxa from the estimated community.

For monitoring applications, the two directions imply different responses. If OSCD$^{+}$ dominates, the system may need stronger rejection of unknown-like samples or improved handling of non-target particles. If OSCD$^{-}$ dominates, the threshold is likely too strict for abundance estimation, and known taxa are being undercounted. Reporting only a single community distance would obscure this distinction. These directional diagnostics therefore provide a bridge between machine-learning errors and ecological interpretation.

\subsection{Deployment assumptions and limitations of community-aware calibration}
\label{subsec:discussion-ca-limitations}

Community-aware threshold calibration improved OSCD when validation communities represented the deployment condition and when the fixed-recall baseline left room for improvement. Its limitations followed the same logic. Empirical validation communities on SYKE-ZooScan 2024 had a much lower unknown ratio than empirical test communities, non-target-enriched settings were difficult when non-target unknowns were absent from positive-ratio validation communities, and ZooLake left little MSP gain because fixed known recall was already close to the community-aware threshold.

Another limitation is the choice of community objective. OSCD is appropriate when the target is relative community composition over known taxa plus an unknown bin, but it is not a universal surrogate for all ecological summaries. Diversity, richness, and dominant-taxon metrics may require target-specific calibration, especially when rare taxa or dominance structure are the monitoring priority.

Finally, the experiments are based on a frozen DINOv2 encoder and post-hoc OOD scores. This design isolates the evaluation and threshold-selection problem, but it does not exhaust the space of possible open-set models. Fine-tuned encoders, open-set training losses, generative unknown models, or taxonomic hierarchy-aware classifiers may improve sample-level separation or alter the relative performance of the scores tested here. The community-level threshold-selection issue would still need to be evaluated for those alternatives.

\subsection{Implications for automated plankton monitoring}
\label{subsec:discussion-monitoring}

The results suggest several practical guidelines for automated plankton monitoring systems. First, detection F1, Youden index, and FPR@95\% unknown-recall criteria should not be used as deployment thresholds without checking community-level consequences.

Second, a conservative known-recall threshold can be a reasonable baseline when the expected unknown ratio is low or when validation data are insufficient for representative community calibration. As the expected unknown ratio increases, validation communities become more important for choosing a threshold that preserves abundance estimates.

Third, validation communities should be designed to match the expected deployment condition. If the deployment environment is likely to contain many non-target particles, the validation community set must include comparable non-target structure. If the deployment environment is dominated by a few taxa, dominant-taxa validation scenarios should be included. If rare taxa and diversity are the monitoring priority, the calibration objective should include diversity-sensitive metrics rather than OSCD alone.

More broadly, open-set plankton recognition should be evaluated as a component of an ecological measurement pipeline, not only as an image-level detector. The relevant question is not merely whether unknown samples are rejected, but how the rejection rule changes inferred community composition. Community-level metrics and directional error diagnostics provide a way to make that question explicit.

\section{Conclusion}
\label{sec:conclusion}

Automated plankton image recognition should be evaluated as an abundance-estimation problem, not only as an image-level detection problem. This study examined that deployment setting using three datasets, controlled pseudo-communities, post-hoc OOD scores, and threshold-selection strategies that connect sample-level decisions with ecological abundance error.

OSCD and its directional components show whether community error arises from unknown absorption into known taxa or from over-rejection of real known taxa into the unknown bin. This distinction mattered empirically: unknown images were frequently absorbed with high confidence, but aggressive sample-level thresholds often produced larger errors by underestimating known taxa.

Community-aware threshold calibration improved OSCD on two of three datasets and matched the fixed-recall baseline on the third. Its benefit is therefore conditional: it requires validation communities that reflect deployment and a meaningful threshold trade-off in the OOD score. It does not replace conservative known-recall baselines when those conditions are not met. When diversity, richness, or dominance are the monitoring priority, OSCD-based calibration should also be paired with endpoint-specific objectives. Future work should evaluate real cross-system deployments, quantify validation-test community representativeness, and combine community-level evaluation with open-set training or fine-tuning methods.

\section*{Code and data availability}

The analysis code, pseudo-community generation scripts, threshold scans, multi-seed robustness scripts, and figure-generation scripts will be deposited in a public repository at publication. The experiments use publicly available datasets: SYKE-ZooScan 2024 \citep{kareinen2024opensetplankton}, SYKE-IFCB 2022 \citep{kraft2022operational}, and ZooLake \citep{merz2021aquascope}. Processed split files, pseudo-community manifests, and derived metric tables will be released with the code to support reproducibility, subject to the redistribution terms of the original datasets.

\bibliographystyle{cas-model2-names}
\bibliography{references}

\end{document}